\pdfoutput=1

\documentclass[11pt]{article}

\usepackage{authblk}

\usepackage[]{naacl2021}

\usepackage{times}
\usepackage{latexsym}

\usepackage{hyperref}
\usepackage{url}
\usepackage{caption}
\usepackage{subcaption}
\usepackage{graphicx}
\usepackage{multirow}
\usepackage{makecell}
\usepackage{booktabs, siunitx}
\usepackage{comment}

\graphicspath{{picture/}}

\usepackage[T1]{fontenc}

\usepackage[utf8]{inputenc}

\usepackage{microtype}

\title{Put Chatbot into Its Interlocutor's Shoes: \\
New Framework to Learn Chatbot Responding with Intention}

\author[1*]{Hsuan Su}
\author[2*]{Jiun-Hao Jhan}
\author[2$\dagger$]{Fan-yun Sun}
\author[2$\Diamond$]{Saurav Sahay}
\author[1*]{Hung-yi Lee}
\affil[*]{National Taiwan University, Taipei, Taiwan}
\affil[$\dagger$]{Stanford University, Stanford, CA, USA}
\affil[$\Diamond$]{Intel Labs, Santa Clara, CA, USA}
\affil[ ]{\textit {\{b04203058, r06942141, hungyilee\}@ntu.edu.tw}}
\affil[ ]{\textit {fanyun@stanford.edu}}
\affil[ ]{\textit {saurav.sahay@intel.com}}

\begin{document}
\maketitle
\begin{abstract}

Most chatbot literature that focuses on improving the fluency and coherence of a chatbot, is dedicated to making chatbots more human-like. However, very little work delves into what really separates humans from chatbots -- 
humans intrinsically understand the effect their responses have on the interlocutor and often respond with an intention such as proposing an optimistic view to make the interlocutor feel better. This paper proposes an innovative framework to train chatbots to possess human-like intentions. Our framework includes a guiding chatbot and an interlocutor model that plays the role of humans. The guiding chatbot is assigned an intention and learns to induce the interlocutor to reply with responses matching the intention, for example, long responses, joyful responses, responses with specific words, etc. We examined our framework using three experimental setups and evaluated the guiding chatbot with four different metrics to demonstrate flexibility and performance advantages. Additionally, we performed trials with human interlocutors to substantiate the guiding chatbot's effectiveness in influencing the responses of humans to a certain extent. Code will be made available to the public.
\end{abstract}

\section{Introduction}

Humans have evolved to become sensitive to their social interactions. The more they interact, the more they generally learn what to say and what not to say to light up people's mood or to avoid upsetting others. In this paper, we aimed to train a chatbot to emulate these human-like qualities by making it learn from interactive conversation. A chatbot that understands the effect its utterances have on the interlocutor could be a significant step towards achieving human-level chatbots. 

\begin{figure}[t]
    \centering
    \includegraphics[width=\linewidth]{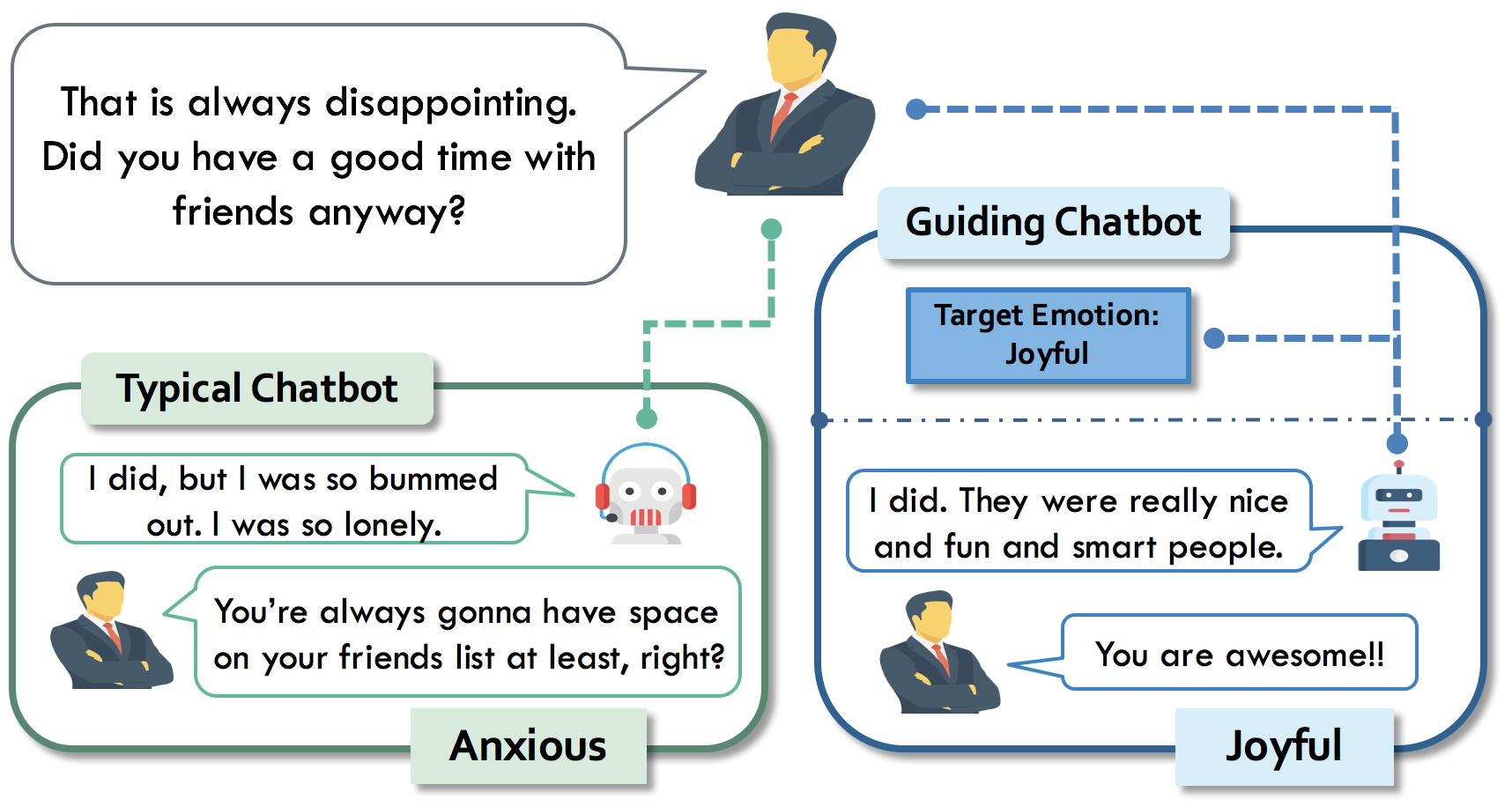}
    \caption{An example of the dialogue to show that how the chatbot interacts with the interlocutor, and how our chatbot affects the interlocutor's response when assigned the intention, \textit{making people respond joyful.}
    }
    \label{fig:intro}
\end{figure}

A chatbot that understands the effect of its utterances on the interlocutor is also critical in real-world applications. 
For instance, as shown in Figure \ref{fig:intro}, given a context from the interlocutor, both responses "I did. They were really nice and fun and smart people." and "I did. I was so bummed out. I was so lonely." were relevant and reasonable responses, and were equally suitable for a typical chatbot.
However, we could give an intention to the proposed chatbot (guiding chatbot), such as making the interlocutor feel joyful.
In this way, the chatbot would respond in a positive way to induce joy in the interlocutor. 

Much literature combine Reinforcement Learning(RL) \citep{kaelbling1996reinforcement} with transformer-based \citep{NIPS2017_3f5ee243} models to control the chatbot's output. \citet{gupta-etal-2019-writerforcing} proposed models to concentrate on crucial keyphrases presented in the context. Their models tended to generate outputs that were more coherent and specific to the conditionals, which leaded to more non-generic words. By training with a combination of the above criteria, their approach leaded to more diverse and interesting responses. However, these previous works focused on controlling the chatbot's responses and completely neglected the interlocutor in their training.

In this paper, we made extensive use of the interlocutor's responses as interactive experiences to train our guiding chatbot to influence the interlocutor with intentions. 
We introduce a novel training framework, in which there were two conversational models that simulated chatbot-interlocutor interaction. One model acted as the interlocutor model, while the other was the guiding chatbot to be trained. 
The interlocutor took the guiding chatbot's output as its input and generated corresponding responses. 
The guiding chatbot was given a controllable factor, which represented the intention it had.
We defined reward functions according to three different controllable factors, sentence length, emotion, and specific words, to make the guiding chatbot learn to induce the interlocutor model to generate desired responses using RL. 

To evaluate our guiding chatbot, we designed several experiments to examine the rewards corresponding to three controllable factors, and empirical results demonstrate that our guiding chatbot can influence humans' responses. Moreover, we found that training with more interlocutor models together improved the guiding chatbot's performance on the human evaluation experiment.
Furthermore, we analyzed recent off-the-shelf chatbots based on experimental results, aiming to find hidden tendencies these chatbot models had, such as cursing more or being more irritative.

\section{Related Work}
The most common chatbot model is sequence-to-sequence based \citep{sutskever2014sequence}. Recently, numerous researchers applied transformer to build coherent chatbots by retrieval-based \citep{zhou-etal-2016-multi, wu-etal-2019-sequential, yang-etal-2018-learning, henderson2017efficient, Yan2016LRDNN} and generative-based \citep{ritter-etal-2011-data, DBLP:conf/aaai/SerbanSBCP16, shang-etal-2015-neural, DBLP:conf/aaai/TammewarPJNM18} approaches. Despite the decent fluency and coherence these chatbots achieved, they still hardly converse like a human. The reason might be that they are essentially devoid of emotions.

Furthermore, some used RL to improve their chatbot's performance \citep{DBLP:conf/aaai/SerbanSBCP16,williams1992simple} and others combined RL with GPT-2 \citep{radford2019language} models to control the sentiment of their chatbot's response to make it more user-friendly \citep{han2019adversarial, lee2018scalable}. Beyond the viewpoint of sentiment, the EmpatheticDialogues (ED) dataset was collected \citep{rashkin-etal-2019-towards} to train a chatbot that could recognize the feeling of the interlocutor and know how to reply accordingly \citep{Lin_Xu_Winata_Siddique_Liu_Shin_Fung_2020}. However, these researchers neglected what really separated humans from chatbots -- humans understand the impact their responses have on the interlocutor and 
often responded with intentions and expectations. Note that it is not just about being empathetic as human's intentions could vary widely.

One previous work also considered interlocutor responses \citep{shin2019happybot}. It used a sentiment predictor to predict the interlocutor's sentiment given the chatbot's response, and also trained the chatbot with RL. Unlike this previous work, our proposed framework explicitly modeled the possible responses of interlocutors. Explicitly modeling interlocutor responses give the proposed framework more flexibility. For example, in addition to steering the interlocutor's sentiment as in this paper, the framework could be used to build a chatbot that induce the interlocutor to become more talkative by setting its learning target to be making the interlocutor generate longer sentences. Moreover, we also developed techniques to preserve the dialogue's coherence, so our chatbot could still generate fluent and appropriate responses in addition to having a particular intention.

Apart from influencing the interlocutor, the proposed framework also served as a way to analyze the underlying inclination of the off-the-shelf chatbots playing the role of interlocutor.
Through the interaction, we could know what factors are apt to influence these off-the-shelf chatbots. 
\citet{holzinger2017need} claimed that the appealing performance of recent robust and SOTA models belied a potential problem of black-box models: these models lacked an explicit declarative knowledge representation. Hence, calling for a transparent representation, they dug into explaining trained models. In contrast to the previous contributions, we tried to explain the implied tendency of a chatbot, which was not obvious to recognize. According to the experiments, we were capable of telling whether the off-the-shelf black box chatbot possessed certain predispositions, such as tending to swear more or having a short temper.

\section{Methodology}
\subsection{Framework}
\begin{figure*}[htp!]
    \centering
    \includegraphics[width=\linewidth]{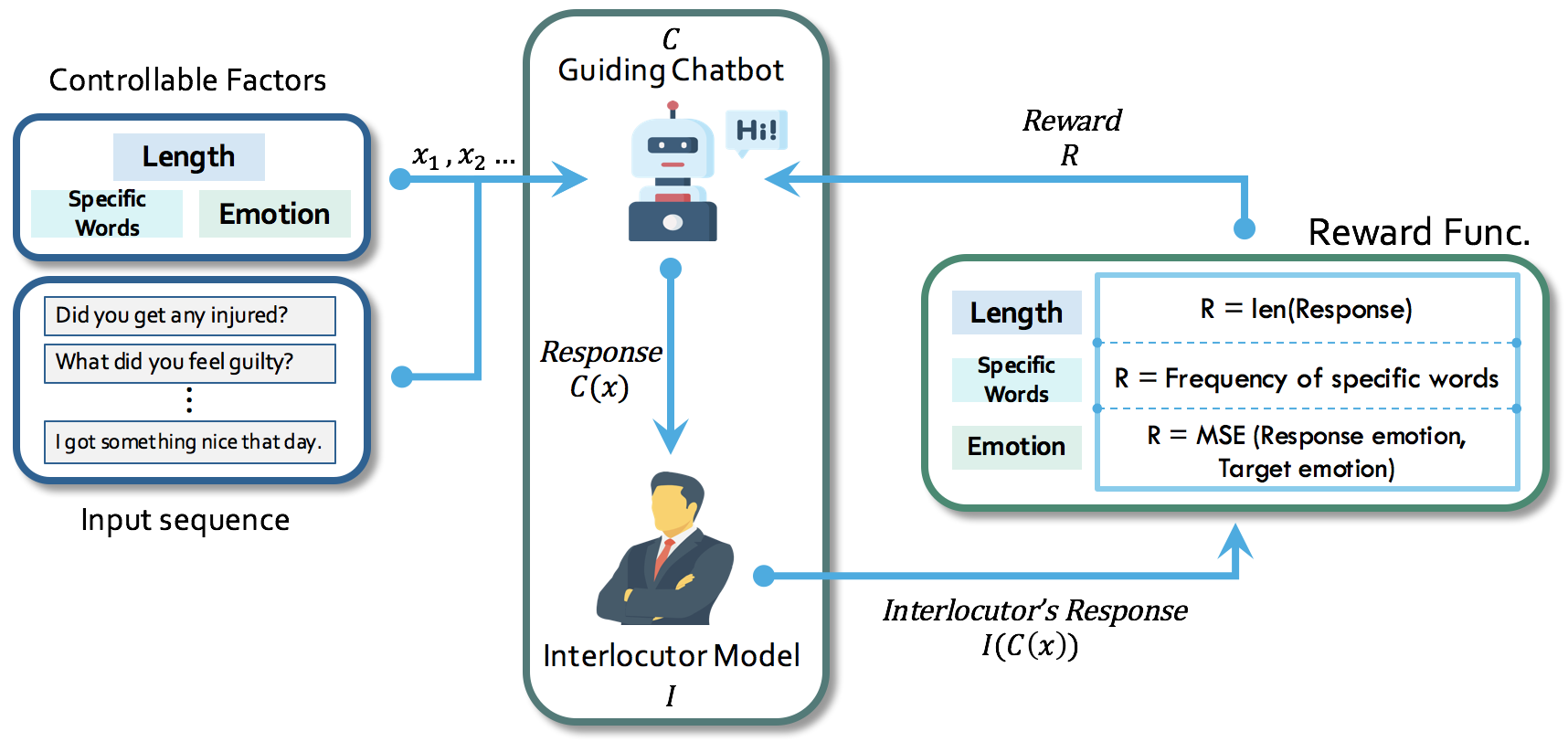}
    \caption{The framework that we proposed to teach the guiding chatbot how to achieve the intention assigned by the controllable factors.}
    \label{fig:framework}
\end{figure*}


The proposed framework is shown in Figure \ref{fig:framework}. It consisted of two conversational models: the guiding chatbot and the interlocutor model.
The interlocutor and guiding chatbot simulated the dialogue between a human and a chatbot.
The guiding chatbot aimed to generate a response that maximize rewards according to different controllable factors; the interlocutor models produced responses based on the guiding chatbot's response in order to simulate a human's response. Therefore, grounded in different controllable factors, we examined corresponding rewards to optimize the guiding chatbot to influence the interlocutor. 

\subsection{Conversational Models}

\paragraph{Interlocutor}
The model $I$ represented the interlocutor. 
$I$ could be any off-the-shelf chatbot whose parameters were fixed during  training; that is, it was unnecessary to know its parameters in the framework. 
$I$ was only used during the training phase to train the guiding chatbot via interaction. In the testing phase, the guiding chatbot will interact with real human beings. 
The interlocutor models’ settings will be described in Section~\ref{sec:model_set}.

\paragraph{Guiding Chatbot}
The guiding chatbot model C was the chatbot we trained to induce desired responses in the interlocutor. 
We built the guiding chatbot model $C$ based on DialoGPT \cite{zhang-etal-2020-dialogpt}.
To train model $C$, given the input sentence $x$, our chatbot $C$ generated a sentence $C(x)$.
The generated sentence $C(x)$ then became the input for $I$, and $I$ output its response $I(C(x))$. 
We defined the reward $R$ for $C$ based on $C(x)$ and $I(C(x))$, and $C$ was trained to maximize the value of $R$ by the policy gradient.
The definition of the reward $R$ depended on the controllable factors, that is, the intention of the guiding chatbot (how the guiding chatbots wanted the interlocutor to respond).
The definition of our reward functions is in Section~\ref{sec:reward}, and the controllable factors are in Section~\ref{sec:control_factors}.

\subsection{Rewards Functions}
\label{sec:reward}
We introduce two kinds of reward functions: intention reward $R_I$ and coherence reward $R_C$. The final reward that the guiding chatbot $C$ learned to maximize will be a combination of $R_I$ and $R_C$.

\paragraph{Intention}
To influence the interlocutor, the guiding chatbot $C$ ought to learn from the interlocutor's reaction. 
To be more specific, we collected responses $I(C(x))$ from the off-the-shelf chatbots when interacting with our guiding chatbot. 
Then the intention reward $R_I$ was obtained by evaluating the interlocutor's responses, that is, $I(C(x))$, based on the controllable factors of guiding chatbot $C$.
Using the intention reward allowed the guiding chatbot to induce the interlocutor to perform specifically according to the controllable factors, namely our intentions. 
The formulation of $R_I$ depended on the controllable factors.
To observe the effectiveness of guiding these interlocutor models, in this paper, we had three controllable factors, which were equal to our intentions: to extend the sentence length, to make the interlocutor speak with a particular emotion, and to induce the interlocutor to speak specific words.
Exact formulation of rewards for different controllable factors will be given in Section~\ref{sec:control_factors}.
\paragraph{Coherence}
Using the intention reward as the only reward leaded to a drawback that the guiding chatbot ignored the coherence between the input $x$ and the generated response $C(x)$.
To avoid this problem, an extra constraint on the guiding chatbot to maintain coherent responses was necessary: we applied another conversational model $C^\prime$ that served as a constraint maintaining coherence. Here we used the open-domain GPT-2 model as the $C^\prime$. To be more specific, we estimated the difference in generated probability between $C$ and $C^\prime$ and minimized the estimated difference. As a result, $C$ would be less likely to produce responses unrelated to input $x$ coherent to responses generated by $C^\prime$.
The additional reward $R_C$ is defined as below.
\begin{equation}
R_C = P_{C^\prime}\left(C(x) \mid x\right).
\end{equation}
$R_C$ was the likelihood that $C^\prime$ generated the sentence $C(x)$ given the input sentence $x$. 
This term served as a kind of regularization that avoids drift during training.

To sum up, the total reward is defined as: 
\begin{equation}
R = \lambda R_I + (1-\lambda) R_C,
 \label{eq:R}
\end{equation}
where $\lambda$ is the hyper-parameter.

\section{Controllable Factors}
\label{sec:control_factors}
Below are the three types of controllable factors studied in this paper.
$R_I$ in Section~\ref{sec:reward} could be either $R_L$ for sentence length, $R_E$ for emotion, or $R_W$ for specific words, introduced below. 

\paragraph{Sentence Length}
A chatbot that could inspire the interlocutor to become more talkative is desirable in many real world applications. We aimed to observe whether our chatbot was able to make the interlocutor more talkative, and extend the length of conversations. Hence, we counted the sentence length of interlocutor models' responses as $R_L$. By optimizing this reward, we anticipated that the guiding chatbot might extend sentence length from the interlocutor.

\paragraph{Emotion}
We studied whether our chatbot was capable of inducing the interlocutor to respond with different emotions. 
We selected eight emotions, including anger, anxiety, contentment, disgust, hope, joy, sadness, surprise. 
We selected the 8 emotions such that two emotions are located in each of the four different quadrants of the Valence-Arousal coordinate \citep{model_of_affect}. 

To ascertain the emotion of sentences, we established an Emotion Detector. 
The Emotion Detector was an emotion classifier used to classify emotion given an input sentence. 
We trained the Emotion Detector on the EnpatheticDialogue (ED) dataset \citep{rashkin-etal-2019-towards}. For each sentence, the Emotion Detector will employ a Valence-Arousal (VA) projection grounded on the Valence-Arousal coordinate \citep{model_of_affect, VA_guidance}. 
Given an input sequence, the Emotion Detector would output a two-dimensional vector representing sequence's emotion, defined as emotional valence\footnote{e.g. fear=[-0.12, 0.79], joy=[0.85, 0.15]}.
More details related to the Valence-Arousal Coordinate will be discussed in Section~\ref{sec:VA}.
We utilized the BERT \citep{devlin-etal-2019-bert} architecture, a pre-trained contextualized embedding model, to improve language understanding. Next, we fine-tuned the BERT model on an emotional classification task to enhance the model's capability of categorizing each emotion. The accuracy of our emotion detector was up to 82\%, and, therefore, we could obtain a detected emotion and its emotional valence given an input sentence. 

The Emotion Detector takes $I(C(x))$ as input and predicted its emotional valence according to the VA coordinate. Therefore, we could calculate the Mean Square Error (MSE) between the emotional valence of the interlocutor models' responses and the target emotion's emotional valence as the reward $R_E$.

\paragraph{Specific Words}
We aimed to induce the interlocutor to speak with words from specific groups. These word groups, including Bad Words, Sports, and Food, were collected from Google's team\footnote{\url{https://gist.github.com/jamiew/1112488}} and the EnchantedLearning website\footnote{\url{https://www.enchantedlearning.com/home.shtml}}. To provoke the interlocutor to respond to the sentence including the specific words we want, we calculated the frequency of the specific words in a sentence. We counted the frequency of interlocutor models' responses that contain words in these word groups as $R_W$. We anticipated that the interlocutor can generate a sentence that contains more words from the specific group and still be coherent as well as fluent.

\section{Experimental Setup}
\subsection{Dataset}
\paragraph{EmpatheticDialouges Dataset} \citet{rashkin-etal-2019-towards} created an innovative dataset with around 25K conversations, each consisting of a speaker and a listener. The participants, acting as the speaker, initiated the talks, and the psychologists, serving as the listener, responded to the speaker empathetically. The dataset covers 32 different emotion labels including positive, negative, and neutral emotions. They firmly ensure that each emotion in the dataset was evenly distributed. Nonetheless, a few emotion classes were quite similar, such as "sentimental" and "nostalgic". Thus, we merged these equivalent emotion classes into one emotion class.

\subsection{Valence-Arousal Coordinate Projection}
\label{sec:VA}
In Valence-Arousal Coordinate study \citep{model_of_affect, VA_guidance}, researchers assigned emotional values to nineteen kinds of emotions. We performed supervised training of the Emotion Detector based on these known emotions on the ED dataset. Each emotion could be represented as a  two-dimensional vector. Therefore, we could map each emotion to the coordinate on the VA space.

\subsection{Model Settings}
\label{sec:model_set}

\paragraph{RL Training Details}
We applied the Policy gradient \citep{sutton2000policy} as our RL algorithm. To implement an RL training chatbot, we applied the DialoGPT model, which fine-tuned the GPT-2 model on 147M multi-turn dialogues from Reddit discussion threads. The GPT-2 model was a transformer-based model with 36 layers, 20 attention heads in each layer, 345M parameters, and an embedding size was 1024. This model was trained on the WebText dataset and 50,257 tokens with invertible byte pair encoding to preserve capitalization and punctuation. In our training procedure, we fine-tuned the DialoGPT model on the ED dataset based on the reward function mentioned in Section~\ref{sec:reward}.

\paragraph{Interlocutor Models}
 The interlocutor models had three different setups: 
\begin{itemize}
    \item The Publicly available \textbf{Google bot}  \citep{vinyals2015neural}\footnote{\url{https://github.com/Conchylicultor/DeepQA}} was trained on the dataset proposed by  \citet{Danescu-Niculescu-Mizil+Lee:11a} with 220,579 conversational exchanges between 10,292  pairs. The whole corpus was split into training and testing sets.
    \item  The same \textbf{DialoGPT} model mentioned in Section~\ref{sec:model_set} was used here to act as the interlocutor. The weights of the model were fixed.
    \item A BERT-based \textbf{Retrieval} chatbot trained on the ED dataset. Given input sentences, the chatbot chose the corresponding response from the candidate pool. The BERT encoder first embedded the sentences into sentence embedding and then computed cosine similarity between the input sentences and all candidates to select the most likely option. The candidate pool was comprised of all sentences in the ED dataset, which contained approximately 100K sentences.
\end{itemize}




\begin{table*}[]
\centering
\resizebox{\textwidth}{!}{%
\begin{tabular}{cc|cccc|cccc|cccc}
\toprule{}
\multirow{2}{*}{\textbf{\begin{tabular}[c]{@{}c@{}}Interlocutor\\ while training\end{tabular}}} & \multirow{2}{*}{\textbf{\begin{tabular}[c]{@{}c@{}}Interlocutor \\ while testing\end{tabular}}} & \multicolumn{4}{c}{\textbf{Sentence Length}} & \multicolumn{4}{c}{\textbf{Emotion (Anxiety)}} & \multicolumn{4}{c}{\textbf{Specific Words (Food)}} \\ \cline{3-14} 
 &  & $R_L$ $\uparrow$ & CPPL $\downarrow$ & PPL $\downarrow$ & SB-3 $\downarrow$ & $R_E$ $\downarrow$ & CPPL $\downarrow$ & PPL $\downarrow$ & SB-3 $\downarrow$ & $R_W$ $\uparrow$ & CPPL $\downarrow$ & PPL $\downarrow$ & SB-3 $\downarrow$ \\ \specialrule{0.05em}{3pt}{3pt}
\textbf{-} & \textbf{GPT-2} & 7.12 & 50.72 & 40.26 & 0.62 & 1.8 & - & - & - & 0.03 & - & - & - \\
\textbf{GPT-2} & \textbf{GPT-2} & 9.5 & 31.82 & 22.84 & 0.79 & 0.78 & 37.39 & 22.61 & 0.77 & 0.38 & 95.68 & 50.7 & 0.68 \\ 
\specialrule{0.05em}{3pt}{3pt}
\textbf{-} & \textbf{Google} & 3.74 & 49.89 & 39.81 & 0.62 & 1.82 & - & - & - & 0.02 & - & - & - \\
\textbf{Google} & \textbf{Google} & 10.14 & 110.59 & 41.67 & 0.91 & \textbf{0.7} & 26.57 & 10.99 & 0.8 & 0.002 & 97.84 & 48.27 & 0.8 \\ \specialrule{0.05em}{3pt}{3pt}
\textbf{-} & \textbf{Ret} & 12.77 & 50.3 & 39.47 & 0.62 & 1.77 & - & - & - & 0.08 & - & - & - \\
\textbf{Ret} & \textbf{Ret} & \textbf{19.79} & 76.55 & 18.46 & 0.94 & 0.75 & 26.57 & 10.98 & 0.8 & \textbf{1.29} & 69.0 & 35.7 & 0.81 \\ \specialrule{0.05em}{3pt}{3pt}
\multirow{3}{*}{\textbf{\begin{tabular}[c]{@{}c@{}}GPT-2 + Google \\ + Ret\end{tabular}}} & \textbf{GPT-2} & 8.52 & \multirow{3}{*}{39.68} & \multirow{3}{*}{30.2} & \multirow{3}{*}{0.75} & 0.52 & \multirow{3}{*}{39.95} & \multirow{3}{*}{34.05} & \multirow{3}{*}{0.71} & \textbf{0.51} & \multirow{3}{*}{72.4} & \multirow{3}{*}{40.55} & \multirow{3}{*}{0.8} \\
 & \textbf{Google} & 4.31 &  &  &  & \textbf{0.5} &  &  &  & \textbf{0.51} &  &  &  \\
 & \textbf{Ret} & \textbf{14.79} &  &  &  & \textbf{0.5} &  &  &  & 0.45 &  &  &  \\ 
 \specialrule{0.05em}{3pt}{3pt}
\textbf{Google + Ret} & \textbf{GPT-2} & 7.95 & 49.31 & 36.13 & 0.78 & 0.53 & 40.27 & 34.15 & 0.71 & 0.08 & 64.33 & 15.55 & 0.99 \\
\textbf{GPT-2 + Ret} & \textbf{Google} & 5.75 & 59.95 & 21.56 & 0.8 & \textbf{0.49} & 41.65 & 33.36 & 0.73 & 0.00 & 64.18 & 51.8 & 0.99 \\
\textbf{GPT-2 + Google} & \textbf{Ret} & \textbf{14.85} & 44.0 & 37.9 & 0.71 & 0.51 & 40.0 & 35.71 & 0.72 & \textbf{0.12} & 246.34 & 15.6 & 1 \\ 
\bottomrule{}
\end{tabular}
}
\caption{
    Results of metrics and rewards according to different controllable factors. The metrics of Conditional Perplexity (CPPL), Perplexity (PPL), and Self-BLEU3(SB-3) are only examined on the guiding chatbot. Rewards are calculated on the interlocutor models during testing. The baseline performance is tested by the original guiding chatbot, the DialoGPT pre-trained model that has not yet trained with any interlocutor model. Higher scores for $R_L$ and $R_W$ indicate better performance. Lower scores for $R_E$, CPPL, PPL, and SB-3 indicate better performance. The best results are boldfaced.}
\label{tab:results}
\end{table*}


\subsection{Evaluation Metrics}

Aside from the reward scores related to the intentions, we also reported the following three metrics in the experiments.

\paragraph{Conditional Perplexity}
The Conditional Perplexity here was to measure the dialogue coherence between the output sentence and input sentence $ x $. 
The equation is shown below.
\begin{equation}
  CPPL = \prod_{i=1}^{T} \frac{1}{(P(C(x)_i|x))^{1/T}}
\end{equation}
$CPPL$ was the conditional perplexity, which was equal to the inverse of the product of each word's probability in the sentence $C(x)$ given the input sentence $x$. $T$ was the length of the sentence $C(x)$.
\paragraph{Perplexity}
Here we employed the pretrained GPT-2 language model to judge if the output sentence $C(x)$ was an acceptable sentence. The computation of Perplexity \citep{chen1998evaluation} is shown below.
\begin{equation}
  PPL = \prod_{i=1}^{T} \frac{1}{(P(C(x)_i))^{1/T}}
\end{equation}
\paragraph{Self-BLEU}
While BLEU score \citep{papineni2002bleu} is usually used to measure the correctness in machine translation,
Self-BLEU \citep{zhu2018texygen} was used here to measure the diversity of chatbot responses; 
we calculated the average BLEU score between sentences in our testing result as the Self-BLEU score.
\subsection{Human Evaluation Setups}
For human evaluation, we recruited participants online.  
There were 19 participants; most of them were graduate or undergraduate students. 
Each participant was given several conversations, including an opening sentence and a corresponding response. They were asked to try to understand the conversation and provide a response to reply to the conversation. 
Therefore, we were able to collect numerous participants' responses to calculate rewards. 
Moreover, participants were asked to score the relevance of the guiding chatbot's response to the opening sentence.
This task was rated on a Likert scale\citep{likert1932technique}, ranging from 1 to 5: Score 1 means a firm disagreement, Score 3 meant neutral, and Score 5 meant an undoubted approval. 
Finally, we counted rewards from humans' responses corresponding to the methods mentioned in Section~\ref{sec:control_factors}.

\section{Discussion and Analysis}
\subsection{Extending Sentence Length}
The first controllable factor was sentence length. We aimed to guide the interlocutor to say more words in a single sentence. Table \ref{tab:results} reveals that our chatbot possessed the ability to encourage the interlocutor to be more talkative. The guiding chatbot interacted with the Google model while training could induce the interlocutor model to increase its sentence length from 3 to 10 words on average. However, as the sentence length increased, the conditional perplexity rose simultaneously. The result reflected that the guiding chatbot trained with the Google model was forced to generate weird sentences so that the interlocutor model would produce a longer sentence. In contrast, although the guiding chatbot trained with the Retrieval model suffered from the same problem, the conditional perplexity increased  only slightly, from 50.3 to 76.55, and the sentence length was much longer. Still, the high Self-BLEU3 score indicates that our chatbot might encounter a low-diversity problem. Therefore, the guiding chatbot trained with the GPT-2 model was the most desirable and stable chatbot to extend the interlocutor's sentence length.

\begin{figure*}[ht]
     \begin{subfigure}{.475\textwidth}
         \centering
         \captionsetup{justification=centering}
         \includegraphics[width=0.98\linewidth]{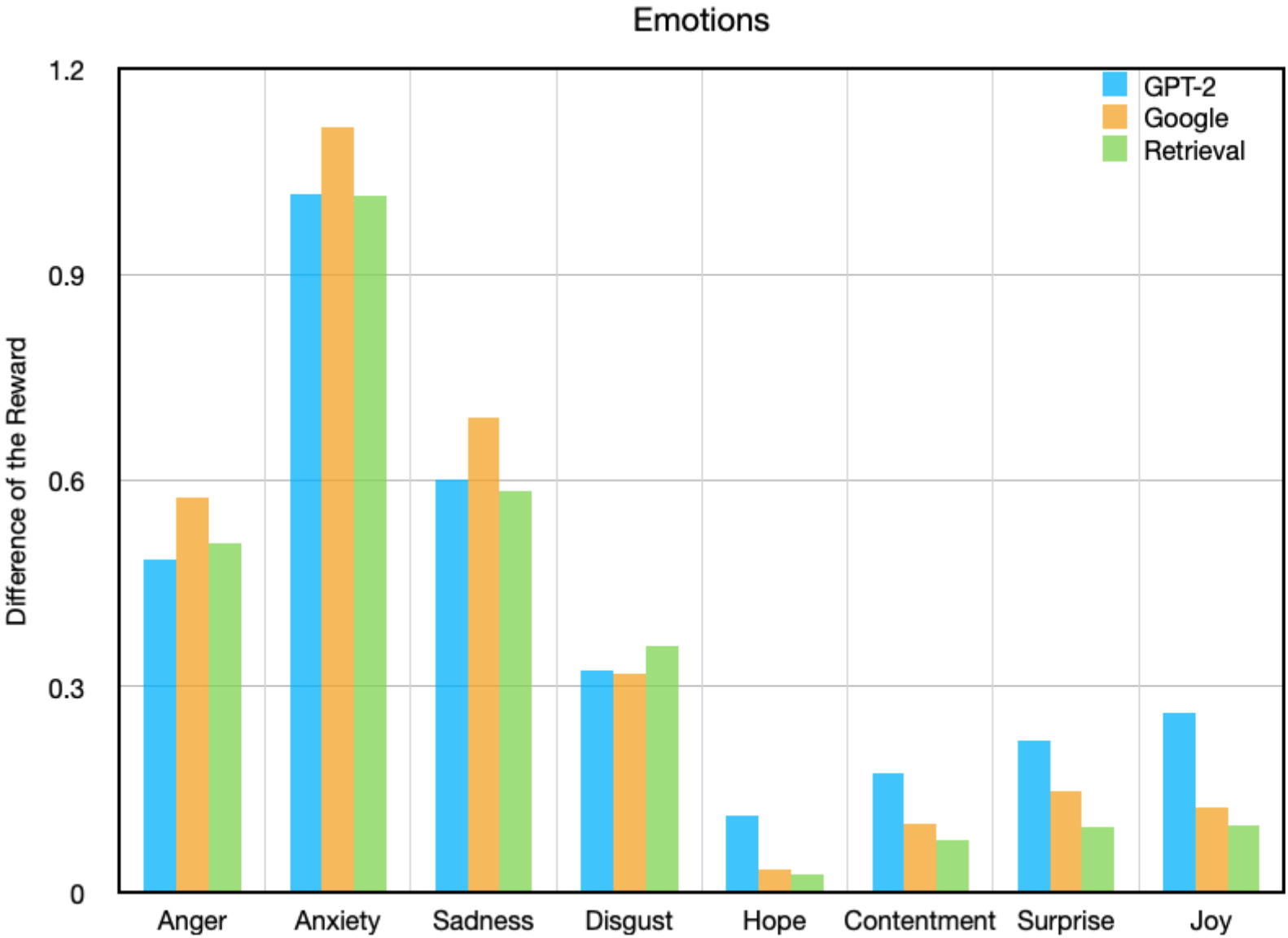}
         \caption{GroundTruth - Reward of Emotions}
         
         \label{fig:gt_reward.png}
         \flushleft
     \end{subfigure}
     \begin{subfigure}{.475\textwidth}
         \centering
         \captionsetup{justification=centering}

         \includegraphics[width=0.98\linewidth]{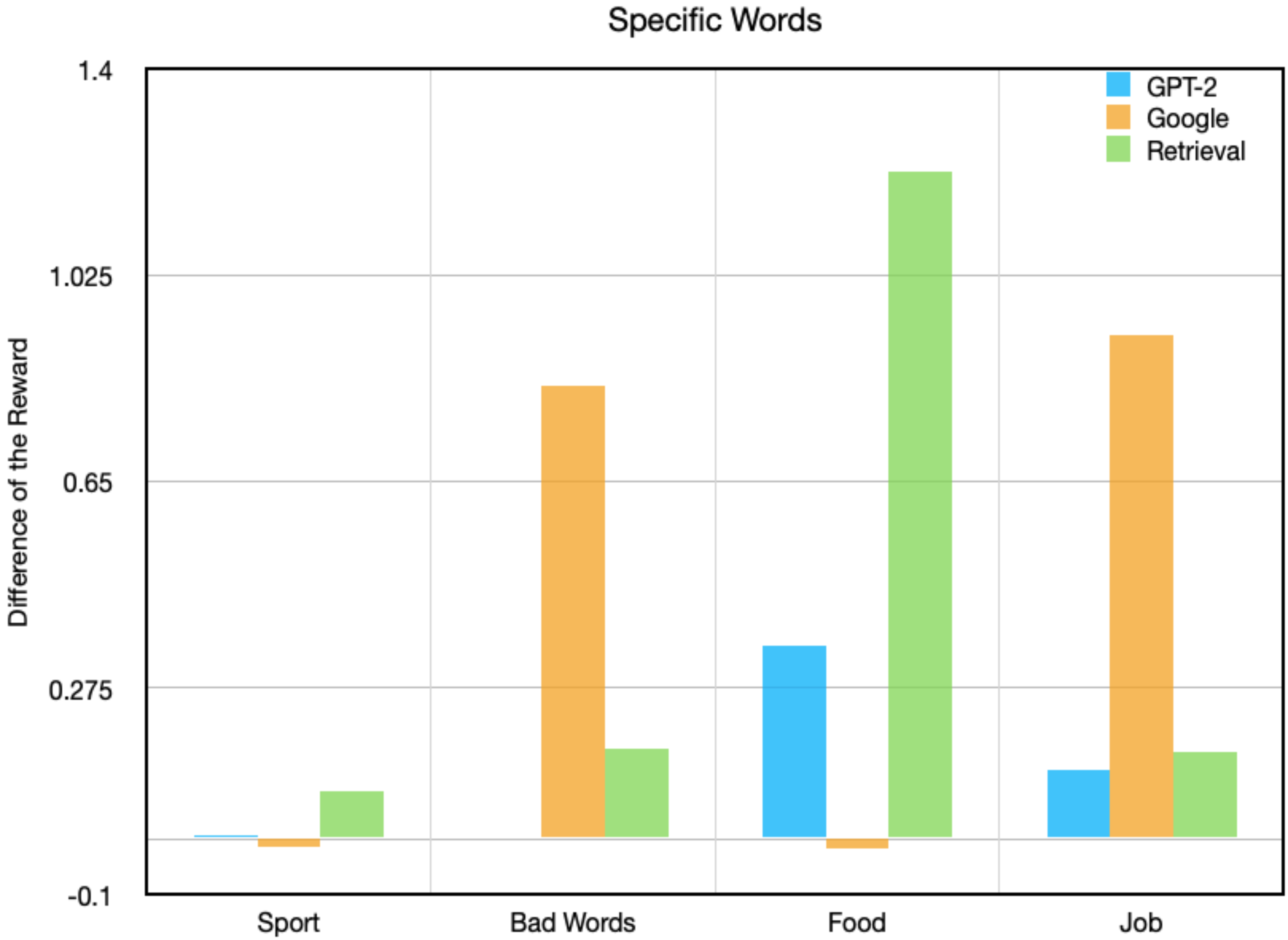}
         \caption{GroundTruth - Reward of Specific Words}
         \label{fig:gt_reward_spword.png}
         
     \end{subfigure}
        \caption{Experiments on controllable factors. The heights of the bars indicate the differences between the rewards of the interlocutor models before and after training.}
        \label{fig:controllable_factors}
\end{figure*}

\subsection{Guiding Emotion}
The second task was to induced the interlocutor to speak with a particular emotion. These emotions included \emph{anger, anxiety, contentment, disgust, hope, joy, sadness, surprise}. We examined the MSE loss between these emotions and the detected emotions of test sentences. Fig. \ref{fig:gt_reward.png} demonstrated that after training, all three interlocutors had similar performance in each emotion. Furthermore, Table \ref{tab:results} indicates that all guiding chatbots trained with any interlocutor model significantly decreased the MSE loss against baseline performance. 
As a result, independent of the choice of interlocutor model, our chatbot could successfully guide the interlocutor to speak with a specific emotion. 

\paragraph{Positive Emotions Versus Negative Emotions}
\begin{table}[]
\begin{tabular}{c|c|c}
\hline
                  & \textbf{Positive} & \textbf{Negative} \\ \hline
\textbf{GPT-2}    & 1.12       & 1.40       \\ \hline
\textbf{Google}   & 1.05       & 1.41        \\ \hline
\textbf{Retrieve} & 1.08       & 1.42       \\ \hline
\end{tabular}
\caption{The MSE scores on the positive and negative emotions of the interlocutors without any fine-tuned.}
\label{tab:emotions}
\end{table}
We investigated how positive/negative of the interlocutors that interacted with our model without any fine-tuning. Table \ref{tab:emotions} shows that all three interlocutors responded more with positive emotions than with negative emotions.

Then, we evaluated how our chatbot realizes the way to influence the interlocutor. Figure \ref{fig:gt_reward.png} shows the difference between the MSE scores of the ground truth sentences and the MSE scores of the test sentences. 
We found that the improvements for negative emotions are greater than those of positive emotions. Table \ref{tab:emotions} shows that the average MSE scores of negative emotions is greater than positive emotions. 
According to the Fig. \ref{fig:gt_reward.png}, the Google model was easier to guided to reply with negative emotions, such as anxiety, sadness, and disgust. In comparison, the GPT-2 model was more easily encouraged to speak with positive emotion, such as joy, surprise, hope, and contentment. 
We attribute this phenomenon to the datasets underpinning each of these chatbots. 
\begin{table*}[]
\centering
\resizebox{\textwidth}{!}{%
\begin{tabular}{c|c|cc|cc|cc}
\toprule{}
\multirow{2}{*}{\textbf{\begin{tabular}[c]{@{}c@{}}Interlocutor\\ while training\end{tabular}}} & \multirow{2}{*}{\textbf{\begin{tabular}[c]{@{}c@{}}Interlocutor\\ while testing\end{tabular}}} & \multicolumn{2}{c}{\textbf{Sentence Length}} & \multicolumn{2}{c}{\textbf{Emotion (Anxiety)}} & \multicolumn{2}{c}{\textbf{Specific Words (Food)}} \\ \cline{3-8}  
 &  & $R_L$ $\uparrow$ & Relevance $\uparrow$ & $R_E$ $\downarrow$ & Relevance $\uparrow$ & $R_W$ $\uparrow$ & Relevance $\uparrow$ \\ 
 \specialrule{0.05em}{3pt}{3pt}
\textbf{-} & \textbf{Human} & 5.82 & 3.10 & 0.41 & 3.10 & 0.05 & 3.10 \\ 
\specialrule{0.05em}{3pt}{3pt}
\textbf{GPT-2} &  & 6.05 & 2.10 & \textbf{0.27} & 3.89 & 0.16 & \textbf{2.63} \\ 
\textbf{Google} & \textbf{Human}  & 2.74 & \textbf{2.31} & 0.47 & \textbf{4.21} & 0.05 & 2.42 \\ 
\textbf{Ret} &  & 5.90 & 1.52 & 0.46 & 3.68 & 0.21 & 1.47 \\ 
\specialrule{0.05em}{3pt}{3pt}
\textbf{\begin{tabular}[c]{@{}c@{}}GPT-2 + Google \\ + Ret\end{tabular}} & \textbf{Human} & \textbf{7.21} & \textbf{2.79} & 0.39 & 3.21 & \textbf{0.68} & 1.53  \\
\hline
\end{tabular}
}
\caption{
Human Evaluation Results. 
Relevance represents the extent to which the guiding chatbot's response is relevant. 
In contrast, the reward is based only on the interlocutor's responses. We tested the baseline performance  of the original guiding chatbot, the DialoGPT pre-trained model that has not yet trained with any interlocutor model. Top results were boldfaced.
}
\label{tab:human}
\end{table*}
The Google model was trained on the Cornell Movie Dialogue dataset, whereas the GPT-2 model was fine-tuned using the ED dataset. 
The movie dataset is full of simple, dramatic, exaggerated sentences.
On the other hand, the ED dataset, designed to arouse the participants' sympathy tends be more positive. 
Furthermore, the Fig. \ref{fig:gt_reward.png} also displays that our chatbot performs exceptionally well on inducing the interlocutor speak with anxiety. The difference in the Google model's reward was up to 0.7, which means that we can significantly induce the interlocutor to speak with anxious emotion.

\subsection{Inducing Specified Words}
 In another set of trials, our chatbot managed to make the interlocutor sentences contain certain groups of words, such as Food, Sports, Jobs, and Bad Words. We calculated the frequency of a word in a specific group. Table \ref{tab:results} shows that the ground truth's reward was close to 0, which suggests that the interlocutor models barely spoke words in the "Food" group before being exposure to by our guiding chatbot. Fig. \ref{fig:gt_reward_spword.png} shows that our chatbot could successfully influence the interlocutor to talk about a sentence containing a word from the "Sports" group and "Food" group. On average, after interacting with the guiding chatbot, the Google model spoke 0.7 more words in the "Job" group, and the Retrieval model was induced to say 0.6 more words in the "Food" group. However, since the rewards of the ground truth are all near 0, Figure \ref{fig:gt_reward_spword.png} indicates that fine-tuning the guiding chatbot using the RL approach can lead the interlocutor to say words they did not previously say.

We also found that the guiding chatbot trained with the GPT-2 model could only weakly induce the interlocutors to use words from the "Bad Word" group. This is almost certainly because bad words rarely appear in the ED dataset. 
The guiding chatbot trained with the Google model was more likely to induce the Google model interlocutor to say words in the "Bad Word" groups. We further analyzed the Cornell Movies dataset and found that, there are 24547 bad words out of 220579 sentences.
We likewise concluded that dramatic utterances in the Cornell Movies dataset brought about the tendency for the interlocutor to say more bad words.

\subsection{Cross Validation of Different Interlocutor Models while Training and Testing}
Having proven that our guiding chatbot can significantly improve all three rewards against ground truth while training with a given interlocutor model, we experimented with the more formidable task of having the guiding chatbot consider all three interlocutor models at once. Table  \ref{tab:results} demonstrates that the guiding chatbot could increase the performance, which indicates that the guiding chatbot could learn more experiences when interacting with more and different interlocutor models. While interacting with more models, the guiding chatbot can improve the "Emotion" and "Specific words" rewards against the guiding chatbot that was only trained with a single interlocutor model. Although the "Sentence Length" reward subtly decreased, the rewards still surpassed the ground truth reward, showing that the guiding chatbot could influence the interlocutor.

Moreover, since we could not assume that our interlocutor models are capable of representing all kinds of humans, we conducted an experiment to evaluate our guiding chatbot all-around. The detailed procedures are as follow: we tested our guiding chatbot on the interlocutor model that our guiding chatbot had seen before during training. For example, the guiding chatbot was trained with the GPT-2 and Google models but would be tested with the Retrieval model. Results in Table \ref{tab:results} shows that all guiding chatbots trained with different interlocutor models could improve the rewards in three controllable factors. Also, we found that while testing on the Retrieval interlocutor model, this model was more likely to be induced to speak longer sentences than other interlocutor models. It is mainly because retrieving a longer response is easier than generating. 

\subsection{Human Evaluation Result}
Human evaluation results sufficiently verify the guiding chatbot's effectiveness of influencing humans' responses to certain extents. Since the performances of the "anxiety" emotion and "Food" group were relatively well, shown in Table \ref{tab:results}, we focused on these factors when conducting the human evaluation. Table \ref{tab:human} shows that the guiding chatbot could significantly induce humans to speak with anxiety, as well as maintain, or even enhance, the relevance within a conversation. This performance was consistent with the results in Table \ref{tab:results}, in which the guiding chatbot acquired the ability to gain better rewards. 

Nonetheless, the results of "Sentence Length" and "Specific Words" can hardly show a promising effect. Although the reward gained improvement slightly, humans generally felt the guiding chatbot's response irrelevant: as the reward increased, the relevance decreased dramatically. 
This result demonstrates that the guiding chatbot might learn a tricky approach to gain higher rewards during training, but this method was not fully adaptive to humans. 
For instance, when training the guiding chatbot to influence the interlocutor to speak the sentence with the "Food" group, the guiding chatbot usually ended up with "What is your favorite food?", ignoring the context. 
In contrast, the guiding chatbot could not only increase $R_{E}$ reward but also improve the coherence between responses of the guiding chatbot and the interlocutor models.
\subsection{Effects of $R_C$}
We analyzed the effects bring by $R_C$. We trained a guiding chatbot model without $R_C$ reward on aforementioned experimental settings in Section~\ref{sec:model_set} and observed that the model was more prone to giving low diversity responses that were irrelevant to the context. In our experiments, the Self-BLEU3 score was near 0.99 and the CPPL was over 10000 without $R_C$ reward.
\section{Conclusion}
This paper introduced a novel framework that aims to train a guiding chatbot to influence the interlocutor. We designed three different controllable factors for the guiding chatbot to induce the interlocutor to reply with responses matching the intention. We managed to prolong the length of the interlocutor's responses, influence the interlocutor to reflect with a particular emotion, and induce the interlocutor to use some specific words more frequently. Furthermore, we further enhanced the performance of the guiding chatbot by training it with more interlocutor models. Experiment results show that our proposed framework can successfully train chatbot with intentions.




\section*{Ethics}
In this paper, we proposed a learning framework that trains chatbots to influence humans. We defined several rewards to reflect different behaviors that we want to induce to humans.

We undertook this work because we envisioned a future in which a chatbot can become a digital companion for humans. To that end, we need the chatbot to be able to understand a human’s mental state and reply with appropriate responses. As a concrete example, chatbots could act as healthcare or relationship coaches for people who could not afford such services. Having a healthcare chatbot to talk to at anytime could alleviate the workload of nurses and therapists.
Moreover, since our framework is reward-agnostic that could be optimize for any reward, we also expect that the experts could customize the profession reward definitions in their fields to bring the technique to higher level usage.

However, we also acknowledge the potential that this technique could be misused. Using our framework, ill-intentioned people could train chatbots with negative intentions and could threaten the stability of our society. 
For example, we have identified the following means by which a malicious actor could take advantage of our proposed technology:

\begin{itemize}
\item \textbf{Emotional Manipulation}: One could train chatbots with the intention of arousing negative emotions such as anxiety, sadness, or anger to influence human’s mental state.
\item \textbf{Social Antagonism}: One could train chatbots with the “Specific Words Intention Reward” to induce the interlocutors to exhibit gender biases or use racist terms to purposefully destabilize society. 
\item \textbf{Political Interference}: One could train chatbots with the malicious intentions of manipulating the public’s political opinion.

\end{itemize}
\hfill \break
To prevent the aforementioned abuse of our method, we propose the following methods to counter them.
\begin{itemize}

    \item \textbf{Intention Classifier}: We could train a dialogue classifier that classifies whether a chatbot is purposefully influencing humans. We believe this is technically achievable as we could find many works that aim to distinguish whether a sentence is generated by humans or not \cite{gao2020dialogrpt}. To further refine this work, we could easily collect training datasets for this classifier by interacting with chatbots trained by our framework and other general-purpose chatbots. By doing this, we could inform humans when we detect that the chatbot they are conversing with is being manipulative.
\item \textbf{Special Token}: In the future, biomimetic technologies could blur the boundary between a living being and an artifact. We suggest that if the chatbot model generates the sentences, the sentence needs to be labeled with some special flag to tell people whether the chatbot generates the sentence with the intention. For instance, we can add  “<chatbot | intention>” before any chatbot’s response with the intention to inform people that a chatbot is trying to influence them. This will make users aware that they are interacting with a chatbot and can undermine the effectiveness of a malevolent attack.  
    \item \textbf{Safety Layer}: Inspired by \cite{adiwardana2020humanlike}, we could use a safety layer (e.g., an additional classifier) to filter out sensitive or toxic responses from chatbots during inference.
\end{itemize}

\paragraph{Future Work} To avoid malicious actors taking our framework and train their own chatbot. The development of the \textbf{Intention Classifier} become an essential research topic. In future work, we would set the development of the Intention Classifier as the top priority. 
The functions of the  Intention Classifier are not only detect the intention of a dialogue system, it can also have an ability to generalize to any other dialogue systems. 
With the power of Meta-Learning~\cite{pmlr-v70-finn17a} the classifier is expected to train on a dialogue system with few data and could have the ability to detect whether sentences generated by the dialogue system are with intention.

\paragraph{}

As developers of emerging technologies, we also take responsibility for defining the boundaries of these technologies. We will continue to refine the aforementioned methods to ensure that the proposed methodology improves public welfare as we intend it to.

\bibliography{anthology,custom}
\bibliographystyle{acl_natbib}

\end{document}